\documentclass[10pt,twocolumn,letterpaper]{article}

\usepackage[final]{wacv}

\usepackage{graphicx}
\usepackage{amsmath}
\usepackage{amssymb}
\usepackage{amsfonts}
\usepackage{bm}
\usepackage{booktabs}
\usepackage{multirow}
\usepackage{array}
\usepackage{xcolor}
\usepackage{tikz}
\usetikzlibrary{arrows.meta,positioning,fit,backgrounds,calc,shapes.geometric}
\usepackage{pgfplots}
\pgfplotsset{compat=1.17}
\usepackage{caption}
\usepackage{subcaption}
\usepackage[skins,breakable]{tcolorbox}

\providecolor{wacvblue}{rgb}{0.21,0.49,0.74}
\definecolor{cmpgray}{gray}{0.55}

\newcommand{\R}{\mathbf{R}}
\newcommand{\rvec}{\mathbf{r}}
\newcommand{\nhat}{\hat{\mathbf{n}}}
\newcommand{\tr}{\operatorname{tr}}
\newcommand{\sgn}{\operatorname{sign}}
\newcommand{\SO}{\mathrm{SO}(3)}
\newcommand{\nts}{\mathrm{nts}}
\newcommand{\tbs}{\mathrm{tbs}}
\definecolor{wacvblue}{rgb}{0.21,0.49,0.74}
\usepackage[pagebackref,breaklinks,colorlinks,allcolors=wacvblue]{hyperref}
\usepackage[capitalize]{cleveref}
\crefname{section}{Sec.}{Secs.}
\Crefname{section}{Section}{Sections}
\Crefname{table}{Table}{Tables}
\crefname{table}{Tab.}{Tabs.}

\def\confName{WACV}
\def\confYear{2027}

\title{Synthetic-to-Real ViT-Based Pose Estimation of a Noncooperative UAV}

\author{
{\large Krishnanujam Srinivas\footnotemark[1]} \quad
{\large Hanish Acharla\footnotemark[1]} \quad
{\large Brij Agrawal} \quad
{\large Leonardo Herrera}\\[8pt]
{\normalsize\scshape Naval Postgraduate School}\\[2pt]
{\small\tt krishnanujam.srinivas@nps.edu, haniacharla@gmail.com, agrawal@nps.edu, leonardo.herrera@nps.edu}
}

\begin{document}
\maketitle
\footnotetext[1]{Work performed as an intern at the Spacecraft Research and Design Center, Department of Mechanical and Aerospace Engineering, Naval Postgraduate School, Monterey.}
\begin{abstract}
Remote pose estimation of noncooperative Unmanned Aerial Vehicles (UAVs) from imagery is critical, as they cannot be influenced or instrumented in advance. Deep-learning-based approaches offer a promising solution; however, their development is constrained by the cost and difficulty of acquiring large-scale real-world datasets with accurate pose labels. Synthetic imagery provides an alternative, but models trained on synthetic data must overcome the synthetic-to-real domain gap to generalize to real-world imagery. This work investigates the inherent synthetic-to-real generalization capability of a Vision Transformer (ViT)-based model for monocular UAV pose estimation. The proposed approach employs a self-supervised DINOv2 backbone and is trained exclusively on labeled synthetic imagery while being evaluated on labeled real-world imagery. Pose ambiguity-aware strategies are incorporated during training and inference to address ambiguities arising from the projection of a three-dimensional target onto a two-dimensional image plane and from target symmetries. An $\alpha$--$\beta$ filter is further integrated during inference to improve pose estimations. To assess the model under operational requirements, it is evaluated in terms of Mean Angular Error (MAE) and inference time, both before and after filtering, using a real-world dataset containing 77,077 labeled UAV images. Before filtering, the model achieves an MAE of $19.18^{o}$ and an inference time of $13.25$ ms, whereas after filtering, these values are $8.74^{o}$ and $13.42$ ms, respectively.

\end{abstract}

\section{Introduction}

Remote pose estimation of Unmanned
Aerial Vehicles (UAVs) from imagery is essential for applications such as autonomous navigation \cite{pessanha2024fixed}, aerial refueling \cite{gong2023research}, and directed-energy systems \cite{herrera2025deep}. The latter application motivates this work, where accurate pose estimation of noncooperative UAVs is critical to automate operational scenarios that are difficult or impractical for human operators. Since noncooperative targets cannot be instrumented or influenced in advance, camera-based remote sensing provides an attractive approach for estimating UAV pose from imagery.

To estimate pose from this imagery, deep learning-based methods have emerged as a powerful solution due to their ability to learn visual representations directly from image data.
Deep-learning-based pose estimation models have shown significant promise in recent years \cite{guan2024survey, ordoumpozanis2025reviewing, liu2026deep}. However, their performance is often limited by the availability of labeled training data. Collecting and annotating large-scale real-world data, especially in the noncooperative UAV context is costly and time-consuming, particularly when accurate pose labels are required. Synthetic data generation provides an attractive alternative by producing large quantities of labeled imagery under controlled variations such as pose, lighting, and viewing conditions. Nevertheless, models trained on synthetic data must overcome the synthetic-to-real domain gap to generalize effectively to real-world imagery.

Similar domain gap challenges have been observed in satellite pose estimation. The second international Satellite Pose Estimation Competition (SPEC2021), co-hosted by the European Space Agency (ESA) and Stanford University's Space Rendezvous Laboratory (SLAB), highlighted the challenge of generalizing models trained on synthetic imagery to realistic imagery \cite{park2023satellite}, motivating research into methods for bridging the synthetic-to-real domain gap. Although focused primarily on satellite pose estimation, the underlying problem is also relevant to noncooperative UAV pose estimation, where acquiring large quantities of accurately labeled real-world pose data is costly and often impractical.

This work investigates whether a Vision Transformer (ViT)-based model can generalize from synthetic to real UAV imagery for pose estimation. The proposed model is based on the self-supervised DINOv2 \cite{oquab2023dinov2} and is trained exclusively on synthetic imagery and evaluated on real-world imagery for direct pose estimation. This experimental setting is intentionally adopted to assess the inherent synthetic-to-real generalization capability of the ViT-based model. Techniques such as adversarial training \cite{ganin2016domain} and domain randomization \cite{tobin2017domain} could be employed to reduce the synthetic-to-real domain gap. However, the primary interest of this study is the inherent synthetic-to-real performance of the ViT-based model without incorporating explicit domain-gap reduction techniques.

ViT leverages self-attention mechanisms to capture global contextual information \cite{dosovitskiy2020image}. ViT has demonstrated strong performance across a wide range of computer vision tasks. However, its effectiveness for synthetic-to-real UAV pose estimation remains insufficiently explored. In this work, the proposed ViT-based model is designed to account for pose ambiguity, a significant challenge in monocular pose estimation arising from the projection of a three-dimensional target onto a two-dimensional image plane and from target symmetries. As illustrated in Figure \ref{fig:ambiguous_poses}, distinct target poses may produce nearly identical image appearances, potentially leading to large pose prediction errors. To mitigate this issue, ambiguity-aware strategies are incorporated during both training and inference.

The proposed model is evaluated using two criteria that are critical for practical deployment: prediction accuracy, quantified by mean angular error, and inference time. In addition, an $\alpha-\beta$ filter is integrated during inference to smooth the estimations. Accuracy and inference time are evaluated before and after filtering to quantify its effect.

The UAV imagery used in this research consists of labeled synthetic and real-world datasets. The synthetic dataset was generated in simulation using a 3D UAV virtual model, whereas the labeled real-world dataset was acquired using a laboratory testbed with 3D printed UAV model, as detailed in \cite{herrera2023deep}. A total of 77,077 labeled real-world images were used in this study.

\section{Preliminaries}
\label{sec:prelim}

This section fixes notation and introduces the four ingredients the method builds on: the rotation representation and error metric (\cref{sec:rot}), the continuous six-dimensional parametrization the network regresses (\cref{sec:sixd}), the four-branch symmetry group that formalizes the pose ambiguity (\cref{sec:ambiguity}), and the radar-cue signal used to resolve it.

\subsection{Rotation representation and error metric}
\label{sec:rot}

A three-dimensional orientation is an element of the special orthogonal group $\SO = \{\R \in \mathbb{R}^{3\times3} : \R^\top\R = I,\ \det\R = +1\}$, the group of proper rotations. $\SO$ is a three-dimensional curved manifold rather than a Euclidean space, and this curvature is the source of both the parametrization difficulties (\cref{sec:sixd}) and the need for a manifold-aware error metric.

Orientations are represented by an axis-angle (Rodrigues) vector $\rvec = \theta\nhat$, where $\nhat$ is the unit rotation axis and $\theta = \|\rvec\|$ is the rotation angle; this is the parametrization in which ground-truth labels are supplied. The corresponding rotation matrix is recovered by the matrix exponential, which for $\SO$ takes the closed form known as the Rodrigues formula,
\begin{equation}
    \R = \cos\theta\, I + (1-\cos\theta)\,\nhat\nhat^\top + \sin\theta\,[\nhat]_\times,
    \label{eq:rodrigues}
\end{equation}
where $[\nhat]_\times$ is the skew-symmetric matrix such that $[\nhat]_\times \mathbf{v} = \nhat \times \mathbf{v}$ for all $\mathbf{v}$. Labels are converted to matrices via \cref{eq:rodrigues} before any loss is computed.

The discrepancy between a predicted and a ground-truth orientation is measured by the \emph{geodesic distance} on $\SO$, the length of the shortest rotation carrying one onto the other, equivalently the angle of the residual rotation $\R_{\text{pred}}^\top \R_{\text{gt}}$:
\begin{equation}
    e(\R_{\text{pred}}, \R_{\text{gt}}) = \arccos\!\left(\frac{\tr\!\left(\R_{\text{pred}}^\top \R_{\text{gt}}\right) - 1}{2}\right).
    \label{eq:geodesic}
\end{equation}
Unlike a Euclidean distance between rotation matrices or quaternions, it reflects true angular displacement and does not saturate or double-count near large rotations. All errors are reported in degrees.

\subsection{Continuous six-dimensional rotation representation}
\label{sec:sixd}

The continuous six-dimensional representation of Zhou~\etal~\cite{zhou2019continuity} is used. The network emits two raw vectors $\mathbf{a}_1, \mathbf{a}_2 \in \mathbb{R}^3$, mapped to a valid rotation matrix by Gram--Schmidt orthonormalization,
\begin{align}
    \mathbf{b}_1 &= \frac{\mathbf{a}_1}{\|\mathbf{a}_1\|}, \notag\\
    \mathbf{b}_2 &= \frac{\mathbf{a}_2 - (\mathbf{b}_1^\top \mathbf{a}_2)\,\mathbf{b}_1}{\left\|\mathbf{a}_2 - (\mathbf{b}_1^\top \mathbf{a}_2)\,\mathbf{b}_1\right\|}, \label{eq:sixd}\\
    \mathbf{b}_3 &= \mathbf{b}_1 \times \mathbf{b}_2, \qquad \R = [\,\mathbf{b}_1\ \ \mathbf{b}_2\ \ \mathbf{b}_3\,]. \notag
\end{align}
The resulting $\R$ is orthonormal with determinant $+1$ by construction, so no post-hoc projection onto $\SO$ is required, and the map is continuous and surjective. This removes the discontinuities that afflict Euler-angle and quaternion outputs and degrade accuracy near their seams.

\subsection{Pose ambiguity and the four-branch symmetry group}
\label{sec:ambiguity}

\begin{figure}[t]
    \centering

    \begin{minipage}[b]{0.47\columnwidth}
        \centering
        \includegraphics[width=\linewidth]{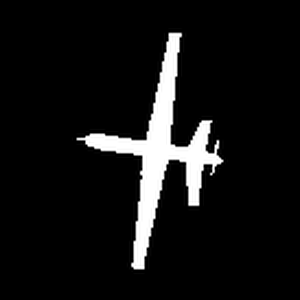}
        
        \footnotesize (a) $\mathcal{R}^{(1)}$ true, $(+,-)$
    \end{minipage}
    \hfill
    \begin{minipage}[b]{0.47\columnwidth}
        \centering
        \includegraphics[width=\linewidth]{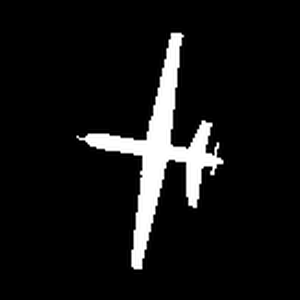}
        
        \footnotesize (b) $\mathcal{R}^{(2)}$ N--T, $(-,+)$
    \end{minipage}

    \vspace{6pt}

    \begin{minipage}[b]{0.47\columnwidth}
        \centering
        \includegraphics[width=\linewidth]{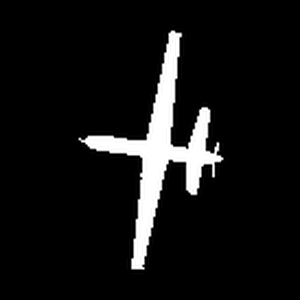}
        
        \footnotesize (c) $\mathcal{R}^{(3)}$ T--B, $(+,+)$
    \end{minipage}
    \hfill
    \begin{minipage}[b]{0.47\columnwidth}
        \centering
        \includegraphics[width=\linewidth]{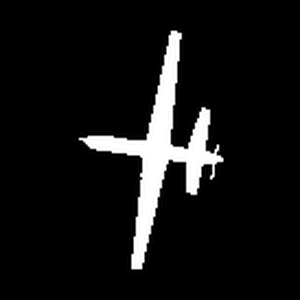}
        
        \footnotesize (d) $\mathcal{R}^{(4)}$ both, $(-,-)$
    \end{minipage}

    \caption{Four orientations of the same aircraft that produce nearly the same silhouette. 
    (a) Tail and top facing the camera; (b) nose and bottom facing the camera; 
    (c) tail and bottom facing the camera; and (d) nose and top facing the camera. 
    The radar-cue bits $(\nts,\tbs)$ beneath each panel differ across all four.}
    \label{fig:ambiguous_poses}
\end{figure}

The defining challenge of this task is that the silhouette of a fixed-wing UAV is approximately invariant under two independent reflections: one exchanging nose and tail, one exchanging top and bottom. As a consequence, a single image is geometrically consistent with four distinct orientations, illustrated in \cref{fig:ambiguous_poses}. Given a true rotation $\R^{(1)}=\R$, these are
\begin{align}
    \R^{(2)} &= \operatorname{diag}(1,1,-1)\,\R\,\operatorname{diag}(1,-1,1), \notag\\
    \R^{(3)} &= \R\,\operatorname{diag}(1,-1,-1), \label{eq:branches}\\
    \R^{(4)} &= \R^{(2)}\,\operatorname{diag}(1,-1,-1), \notag
\end{align}
where $\R^{(2)}$ is the nose--tail flip, $\R^{(3)}$ the top--bottom flip, and $\R^{(4)}$ their composition. Every $\R^{(i)}$ is itself a proper rotation and hence a legitimate regression target: $\operatorname{diag}(1,-1,-1)=\exp(\pi[\mathbf{e}_1]_\times)$ is a rotation by $\pi$ about the body $x$-axis, and the two-sided action in $\R^{(2)}$ composes two reflections of determinant $-1$, one on each side, so each product has determinant $+1$.

The four transforms form a Klein four-group $\mathbb{Z}_2\times\mathbb{Z}_2$ under composition: the identity together with the two reflections and their product, in which each nontrivial element is its own inverse and composing any two distinct flips yields the third. The set $\{\R^{(i)}\}_{i=1}^{4}$ is thus the \emph{orbit} of $\R$ under this group, the complete collection of poses the image cannot distinguish, and a pose in general position has exactly four such preimages. Two properties of this structure are what make the ambiguity tractable rather than merely troublesome. First, the orbit is small and of fixed size: there are precisely four candidates, not an open-ended confusable set, so disambiguation is a two-bit decision. Second, the group acts \emph{faithfully} on a pair of physically measurable sign bits: no nontrivial group element leaves both bits unchanged, so the four orbit members carry four distinct bit patterns and two bits suffice to identify the correct pose, as formalized next.

\paragraph{Radar cues.} To resolve the ambiguity at inference without access to the ground-truth pose, the two \emph{radar-cue} sign bits of~\cite{herrera2025} are adopted, computed from a rotation matrix as
\begin{equation}
    \nts = \sgn\!\big((\R\,\mathbf{e}_1)_z\big), \qquad
    \tbs = \sgn\!\big((\R\,\mathbf{e}_3)_z\big),
    \label{eq:cues}
\end{equation}
with $\mathbf{e}_1=[1,0,0]^\top$, $\mathbf{e}_3=[0,0,1]^\top$, and the convention $\sgn(0)\mapsto+1$. The nose--tail sign $\nts$ is the sign of the $z$-component of the rotated body $x$-axis and encodes which end of the fuselage points toward the sensor. Physically this is the sign of the radar range rate, positive when the nose approaches and negative when the tail does. The top--bottom sign $\tbs$ is the sign of the $z$-component of the rotated body $z$-axis and encodes whether the target is upright or inverted. Each of the three reflections in \cref{eq:branches} flips a different combination of the two bits, so the four orbit members carry four distinct bit patterns (\cref{fig:ambiguous_poses}) and two physically measurable bits identify the correct pose among the four.
\section{Method}
\label{sec:method}

\subsection{Model Architecture}
\label{sec:arch}

\begin{figure*}[t]
    \centering
    \resizebox{\textwidth}{!}{%
    \begin{tikzpicture}[
        font=\normalsize,
        box/.style   ={draw, thick, rounded corners=3pt, align=center, minimum height=1.0cm},
        io/.style    ={box, fill=black!3},
        bbs/.style   ={box, fill=wacvblue!15, draw=wacvblue},
        tok/.style   ={draw, thick, align=center, fill=black!2},
        dec/.style   ={box, fill=black!5},
        cueb/.style  ={box, fill=yellow!22, draw=olive!65!black},
        head/.style  ={box, fill=green!10, minimum height=0.85cm, minimum width=2.0cm},
        loss/.style  ={box, dashed, fill=red!7, draw=red!60},
        tf/.style    ={box, dashed, fill=red!5, draw=red!55, minimum height=0.9cm},
        infer/.style ={box, fill=orange!14, draw=orange!80},
        arr/.style   ={-{Latex[length=2mm]}, thick},
        marr/.style  ={-{Latex[length=2mm]}, thick, black!45},
        larr/.style  ={-{Latex[length=1.7mm]}, red!55, thick, dashed},
        oarr/.style  ={-{Latex[length=2mm]}, orange!80, thick},
    ]

    \node[io, minimum width=1.5cm, minimum height=1.1cm] (img) at (0.75,0) {Input\\image};
    \node[bbs, minimum width=2cm, minimum height=1.4cm] (bb) at (3.20,0)
          {\textbf{DINOv2-L}\\{\small 24 blocks}\\{\small $D{=}1024$}};
    \node[tok, minimum width=2.3cm, minimum height=1.4cm] (tk) at (6.05,0)
          {Tokens\\[1pt]{\small$\mathbf{Z}\in\mathbb{R}^{B\times576\times1024}$}\\[1pt]{\small $+$ LayerNorm}};
    \node[dec, minimum width=1.8cm, minimum height=1.95cm] (s1) at (8.80,0)
          {Stage 1\\decoder\\[3pt]{\small 3 reliable}\\{\small queries}};
    \node[box, fill=green!10, minimum width=1.9cm, minimum height=1.55cm] (kpts) at (11.35,0)
          {Predicted\\keypoints\\[2pt]{\small nose, tail, fin}};
    \node[cueb, minimum width=2.4cm, minimum height=1.95cm] (cue) at (14.20,0)
          {Geometric cue\\$\mathbf{c}\in\mathbb{R}^{6}$ (Eq.~\ref{eq:cue})\\[3pt]{\small deterministic,}\\{\small no parameters}};
    \node[dec, minimum width=1.85cm, minimum height=1.95cm] (s2) at (17.025,0)
          {Stage 2\\decoder\\[3pt]{\small cue-injected}\\{\small queries}};

    \node[head] (kp)  at (19.65, 1.4) {keypoint};
    \node[head] (chi) at (19.65, 0.2) {chirality};
    \node[head] (rot) at (19.65,-1.0) {6D rotation\\{\footnotesize (Eq.~\ref{eq:sixd})}};

    \node[loss, minimum width=3.2cm, minimum height=3.45cm] (ls) at (23.00,0.2)
          {Training objective\\{\small (Eq.~\ref{eq:symloss})}\\[4pt]
           {\small 4-branch geodesic}\\{\small $+\;2.0\times$ keypoint}\\{\small $+\;0.1\times$ chirality BCE}};
    \node[tf] (tfb) at (14.20,-2.85)
          {GT-cue teacher forcing\\{\small $\rho:1.0\rightarrow0.25$}};

    \node[infer, minimum width=2.2cm, minimum height=1.35cm] (rad) at (19.65,-3.55)
          {Radar-cue\\branch select\\{\small (Eq.~\ref{eq:select})}};
    \node[io, minimum width=1.4cm, minimum height=1.0cm] (out) at (22.70,-3.55)
          {$\mathbf{R}_{\mathrm{pred}}$};

    \draw[arr] (img) -- (bb);
    \draw[arr] (bb)  -- (tk);
    \draw[arr] (tk)  -- (s1);
    \draw[arr] (s1)  -- (kpts);
    \draw[arr] (kpts) -- (cue);
    \draw[arr] (cue) -- (s2);

    \draw[marr] (tk.north) -- ++(0,1.2) -| (s2.north);
    \node[black!55, font=\small] at (11.54,2.1) {shared patch-token memory};

    \coordinate (jall) at (18.30,0);
    \draw[thick] (s2.east) -- (jall);
    \fill (jall) circle (1.7pt);
    \draw[arr] (jall) |- (kp.west);
    \draw[arr] (jall) |- (chi.west);
    \draw[arr] (jall) |- (rot.west);

    \draw[larr] (kp.east)  -- (kp.east  -| ls.west);
    \draw[larr] (chi.east) -- (chi.east -| ls.west);
    \draw[larr] (rot.east) -- (rot.east -| ls.west);
    \draw[larr] (tfb.north) -- (cue.south);

    \draw[oarr] (rot.south) -- (rad.north);
    \draw[oarr] (rad.east) -- (out.west);

    \begin{scope}[on background layer]
      \node[draw=black!35, dashed, rounded corners, fit=(kp)(chi)(rot), inner sep=6pt] (hgrp) {};
    \end{scope}
    \node[black!55, font=\small, above=1pt of hgrp] {prediction heads};
    \node[red!60, font=\footnotesize\itshape, below=2pt of ls] {training only};
    \node[orange!80!black, font=\footnotesize\itshape, below=2pt of out] {inference only};


    \end{tikzpicture}
    }
    \caption{\textbf{The cascaded pose decoder.} Dashed red paths are active only during
    training, orange only at inference; everything else runs in both. The per-frame
    pipeline ends at $\mathbf{R}_{\mathrm{pred}}$, which is then passed to the
    sequence-level filter of \cref{fig:abfilter}.}
    \label{fig:arch}
\end{figure*}
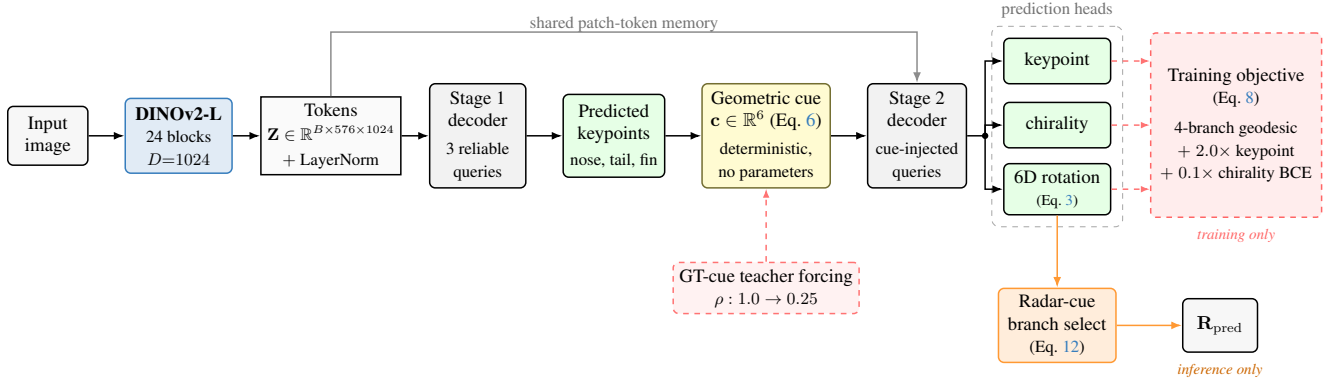

\begin{figure}[tb]
    \centering
    \includegraphics[width=0.78\columnwidth]{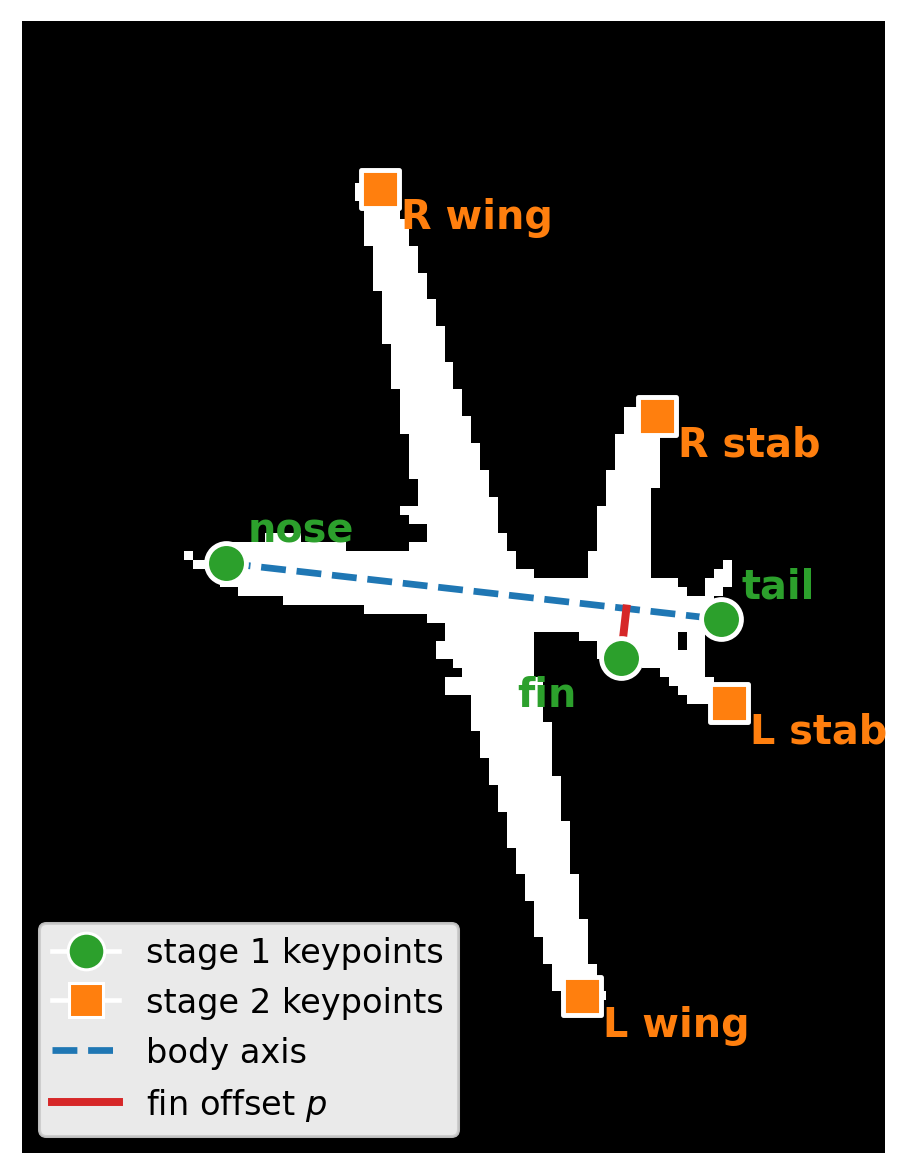}
    \caption{The seven keypoints, shown on a synthetic training sample with its ground-truth labels. Stage~1 (green circles) localizes the three points that are easy to find and stable under a mirror flip: nose, tail, and the fin under the tail. Stage~2 (orange squares) localizes the four left/right-sensitive points: both wingtips and both horizontal tail surfaces. The dashed line is the body axis $\mathbf{u}$ from nose to tail, and the red segment is the fin's perpendicular offset $p$ from that axis, whose sign tells which side of the body the fin is on; these are the quantities that make up the geometric cue of \cref{eq:cue}.}
    \label{fig:keypoints}
\end{figure}

\Cref{fig:arch} presents the model pipeline from left to right. An image crop containing the UAV is first passed through a Vision Transformer backbone, which divides the image into a grid of small patches and produces one feature vector for each patch. These patch features are then used by the rest of the network. The first decoder stage predicts three keypoints that are relatively easy to identify: the nose, the tail, and the small fin beneath the tail. These three points are used to compute a geometric cue that captures the direction of the fuselage in the image and indicates which side of the aircraft the fin lies on. The second decoder stage takes this cue, together with the same patch features, and predicts the four remaining keypoints: the two wingtips and the two horizontal tail surfaces. It also estimates the aircraft's orientation. During training, the full model is supervised by the objective shown on the right side of the figure in dashed red. At inference time, the predicted orientation is passed through the radar-cue correction described in \cref{sec:inference}, shown in orange. This step resolves the remaining nose to tail and top to bottom ambiguities before the final orientation is produced. \Cref{fig:keypoints} shows all seven keypoints on a sample image.

\paragraph{Backbone.} The feature extractor is DINOv2-Large~\cite{oquab2023dinov2}, a ViT-L/14~\cite{dosovitskiy2021vit} with $24$ blocks, hidden width $1024$, $16$ heads, and $\sim\!304$M parameters, pretrained by self-supervised distillation on LVD-142M, a curated collection of 142 million unlabeled images~\cite{oquab2023dinov2}. It was selected for three reasons: its features transfer across visual domains without target-domain labels, which matches the synthetic-to-real setting of this work; it outputs features as a sequence that a query-based decoder can attend to directly; and self-attention lets every patch see the whole aircraft, so that the relationship between, for example, the nose and a far wingtip is available in a single layer. Every input crop, whatever its original size ($200{\times}200$ in the reported experiments; \cref{sec:data}), is resized to $336{\times}336$ before entering the backbone, which divides it into $576$ patch features of width $1024$; after a LayerNorm these form the memory $\mathbf{Z}$ that both decoder stages attend to. All $24$ blocks are fine-tuned, because a frozen encoder's features are nearly identical for an image and its mirror image, and telling those apart is precisely what the model must learn.

\paragraph{Why keypoints, and why in two stages.} Telling the left wing from the right wing in the image is the specific job the keypoints exist to do, and it is worth stating why the network cannot do it directly. DINOv2 is pretrained with augmentations that include horizontal flips, which teach the encoder that an image and its mirror should map to the same representation. Its patch features are therefore close to mirror-invariant by construction, and a single-pass decoder reading them is being asked to name the left wing from evidence that does not distinguish left from right. Fine-tuning pushes the encoder away from that invariance (\cref{sec:impl}), but slowly and incompletely.

Keypoints provide a way around this, because the seven keypoints are not equally affected by a mirror flip. The nose, the tail, and the fin each have no mirror counterpart: each is a single part lying on the aircraft's plane of symmetry, so a mirror flip cannot confuse it with another keypoint. Consistent with this, these three were observed during development to be localized accurately even on images whose orientation output was completely wrong. The wingtips and horizontal tail surfaces, in contrast, come in left/right pairs, and a mirror swaps each with its partner; identifying the left wing is therefore exactly the question the mirror makes ambiguous. Stage~1 accordingly predicts only the three unpaired keypoints. Together they define the body axis, running from nose to tail, and the side of that axis on which the fin appears in the image. That side does flip under a mirror, so it carries the left/right information the backbone features lack. Stage~2 receives it as input and uses it to assign the four paired keypoints consistently with the fin side. This settles which wingtip is which in the image. It does not determine which of the four candidate orientations of \cref{sec:ambiguity} is the true one, because all four present the fin on the same side of the body axis; that choice is left to the radar cue at inference (\cref{sec:inference}). \Cref{fig:keypoints} shows the two groups on a sample image.

Each stage is four transformer decoder layers with eight heads and a small set of learned queries, one per keypoint it must locate, plus a rotation query in stage~2; each query attends to the other queries and to $\mathbf{Z}$, and a small head reads out its image coordinate, or, for the rotation query, the six-dimensional rotation of \cref{eq:sixd} and a left/right (chirality) logit. The keypoint loss supervises all seven keypoints, so stage~1 is trained through the same objective rather than by a separate detector.

\paragraph{The geometric cue (\cref{eq:cue}).} Between the stages, a six-number summary with no learned parameters is computed from the three stage-1 keypoints:
\begin{equation}
\mathbf{c} = \big[\, u_x,\ u_y,\ p,\ a,\ \ell,\ \tanh(25p) \,\big] \in \mathbb{R}^6.
\label{eq:cue}
\end{equation}
All six entries are variables computed per image; none is a fixed constant. Keypoints are expressed in normalized crop coordinates, so each position lies in $[0,1]^2$ and all lengths below are fractions of the crop side rather than pixels. Writing $\mathbf{n}$, $\mathbf{t}$, and $\mathbf{f}$ for the predicted nose, tail, and fin positions, the entries are
\begin{equation}
\begin{aligned}
\ell &= \|\mathbf{t}-\mathbf{n}\|, & \mathbf{u} &= (\mathbf{t}-\mathbf{n})/\ell,\\
p &= u_x (f_y - n_y) - u_y (f_x - n_x), & a &= \mathbf{u}^{\top}(\mathbf{f}-\mathbf{n}).
\end{aligned}
\label{eq:cueterms}
\end{equation}
Here $\mathbf{u}$ is the unit vector pointing from nose to tail, which defines the body axis in the image, and $\ell$ is the axis's projected length. The remaining two entries locate the fin relative to that axis, taking the nose as the origin: $a$ is the fin's distance along the axis, and $p$, the \emph{fin offset}, is its signed perpendicular distance from the axis, positive on one side and negative on the other (\cref{fig:keypoints}). The only constant anywhere in \cref{eq:cue} is the gain $25$ inside the hyperbolic tangent, which is a fixed scaling chosen once and not learned. The sixth entry of $\mathbf{c}$ is a saturated function of the variable $p$ rather than an independent quantity. The vector is defined this way because it captures, in the fewest numbers, three things stage~2 needs and cannot easily get from the features alone. First, $\mathbf{u}$ and $\ell$ give the body's direction and apparent length, so the rotation query starts from a coarse orientation and a foreshortening estimate rather than from nothing. Second, and most important, the \emph{sign} of $p$, whether the fin is to the left or right of the body line, is the one quantity in the image that flips under a mirror while $\mathbf{u}$ does not. It is therefore the single piece of evidence that separates an aircraft from its mirror image, and it is computable from three keypoints the network finds reliably. Because the fin sits very close to the body line, $p$ is a small number; the saturated copy $\tanh(25p)$ passes its sign through at full strength so that stage~2 receives a clear left-or-right signal regardless of how small the offset is. The cue is concatenated onto every stage-2 query through a learned projection, so each query begins its attention already knowing the body axis and the fin side. The benefit is that an ambiguity the backbone features cannot resolve becomes an explicit input: stage~2 can assign the wings \emph{conditioned} on the fin side rather than having to infer the image's handedness from features that are nearly mirror-invariant. Because the cue is a deterministic function of stage-1 output, it adds no parameters and cannot overfit; it only routes information already in the image to the queries that need it. The separate chirality head gives the network a direct, low-dimensional target for the same left/right decision, which stabilizes training of the mirror-sensitive queries. \paragraph{Teacher forcing.} Early in training the stage-1 keypoints are inaccurate, so a cue computed from them is mostly noise, and stage~2 can learn to ignore it entirely. To prevent this, the cue is computed during training from either the ground-truth keypoints or the stage-1 predictions, chosen independently for each sample: with probability $\rho$ the ground truth is used, and otherwise the prediction. The quantity $\rho$ is a sampling probability, not an error measure. It is decayed linearly from $1.0$ to $0.25$ over the first $60\%$ of epochs and held at $0.25$ thereafter, so stage~2 first learns to use a clean cue and then adapts to the noise of the real stage-1 output. At inference the cue always comes from the stage-1 predictions.

\paragraph{Training objective.} Because four different orientations produce nearly the same image (\cref{sec:ambiguity}), a label is really a set of four equally correct answers, and penalizing the network for predicting any one of them would inject error it cannot avoid. The prediction is therefore scored against all four mirror-related candidates of the label and penalize only the distance to the nearest one,
\begin{equation}
    \mathcal{L}_{\text{pose}} = \min_{i \in \{1,2,3,4\}}\, e\!\left(\R_{\text{pred}},\, \R_{\text{gt}}^{(i)}\right),
    \label{eq:symloss}
\end{equation}
with $e(\cdot)$ the geodesic of \cref{eq:geodesic} and $\R_{\text{gt}}^{(i)}$ the four candidates of \cref{eq:branches}. This lets the network commit to whichever candidate it is closest to and be supervised only there. The complete loss adds two auxiliary terms,
\begin{equation}
    \mathcal{L} = \mathcal{L}_{\text{pose}}
    + \lambda_{\text{kp}}\mathcal{L}_{\text{kp}}
    + \lambda_{\text{ch}}\mathcal{L}_{\text{ch}},
    \label{eq:totalloss}
\end{equation}
where $\mathcal{L}_{\text{kp}}$ is a smooth-$L_1$ keypoint loss and $\mathcal{L}_{\text{ch}}$ a binary cross-entropy chirality loss, with $\lambda_{\text{kp}}=2.0$ and $\lambda_{\text{ch}}=0.1$. The terms are kept consistent: the candidate $i^\star$ selected by \cref{eq:symloss} also swaps the left/right keypoint targets when that candidate contains the left--right mirror, so keypoint supervision always refers to the same candidate the pose loss is pulling toward.

\paragraph{Role of the chirality term.} The chirality head predicts a single bit, which side of the body axis the fin lies on, and $\mathcal{L}_{\text{ch}}$ is its binary cross-entropy against the ground-truth side. It is worth being precise about what this term contributes, because the model never uses the predicted bit at inference: the output pose comes from the rotation head, and the branch is fixed by the radar cue of \cref{sec:inference}. The chirality term is a training device, and it helps in two ways. First, it supplies a gradient that the pose loss cannot. By construction \cref{eq:symloss} is indifferent to which of the four mirror-related candidates the network converges to, so it exerts no pressure on the encoder to become mirror-sensitive at all; the chirality loss is the only term in \cref{eq:totalloss} that is minimized solely by distinguishing an image from its mirror, and it is therefore what drives the backbone away from the flip-invariance it inherits from pretraining. Second, it is a far easier target than the pose itself: one bit, read from the same query that produces the rotation, so the shared query representation is pushed toward encoding fin side explicitly rather than implicitly. The term does not make the network select the true candidate orientation; as \cref{tab:ambiguity} shows, that remains the role of the radar cue. Because the bit is unlearnable while the backbone is frozen, the term is disabled in phase~1 and enabled in phase~2 (\cref{sec:impl}).

The two-phase schedule under which the model is trained is given in \cref{sec:impl}.

\subsection{Dataset Augmentation}
\label{sec:sim2real}

\begin{figure*}[t]
    \centering
    \begin{tikzpicture}[
        font=\small,
        img/.style={inner sep=0pt, outer sep=0pt, draw=black!35, line width=0.35pt},
        capstyle/.style={font=\small, align=center, text width=3.1cm},
        subcapstyle/.style={font=\footnotesize, align=center, text width=3.1cm, text=black!55},
        procarrow/.style={-{Latex[length=2.2mm, width=1.6mm]}, line width=0.9pt, wacvblue, shorten >=1.5pt, shorten <=1.5pt},
        cmparrow/.style={{Latex[length=1.6mm, width=1.2mm]}-{Latex[length=1.6mm, width=1.2mm]}, line width=0.6pt, cmpgray, densely dashed, shorten >=1.5pt, shorten <=1.5pt},
        proclabel/.style={font=\footnotesize\itshape, align=center, text width=2.3cm, midway, anchor=south, yshift=4pt, text=wacvblue},
        cmplabel/.style={font=\footnotesize\itshape, align=center, text width=1.9cm, midway, anchor=south, yshift=3pt, text=cmpgray},
    ]
        \node[img] (raw) {\includegraphics[width=3.1cm]{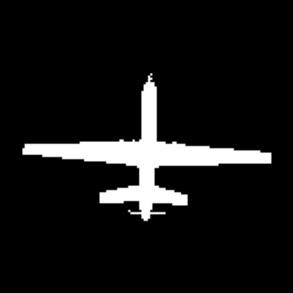}};
        \node[img, right=2.6cm of raw] (aug) {\includegraphics[width=3.1cm]{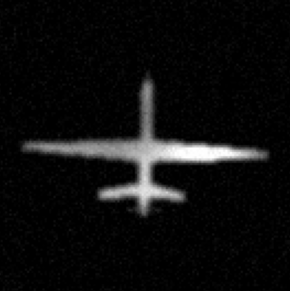}};
        \node[img, right=2.2cm of aug] (real) {\includegraphics[width=3.1cm]{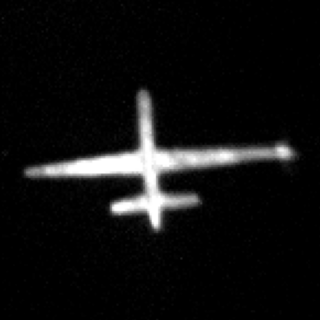}};

        \node[capstyle, below=6pt of raw] (cap-raw) {(a) Raw synthetic image};
        \node[subcapstyle, below=1pt of cap-raw] {binary silhouette};

        \node[capstyle, below=6pt of aug] (cap-aug) {(b) Augmented image};
        \node[subcapstyle, below=1pt of cap-aug] {\cref{sec:sim2real}, applied to (a)};

        \node[capstyle, below=6pt of real] (cap-real) {(c) Real SWIR image};
        \node[subcapstyle, below=1pt of cap-real] {quasi-matched pose};

        \draw[procarrow] (raw.east) -- node[proclabel] {Run Augmentation} (aug.west);
        \draw[cmparrow] (aug.east) -- node[cmplabel] {compare to real} (real.west);
    \end{tikzpicture}
    \caption{Qualitative effect of the calibrated appearance model. (a) A raw synthetic silhouette mask, before any appearance augmentation. (b) The same mask after the calibrated model of \cref{sec:sim2real} is applied, reproducing the soft, non-uniform interior, edge halo, and low but nonzero background of a real sensor image. (c) A real SWIR image at a quasi-matched pose, shown for comparison.}
    \label{fig:augpanel}
\end{figure*}
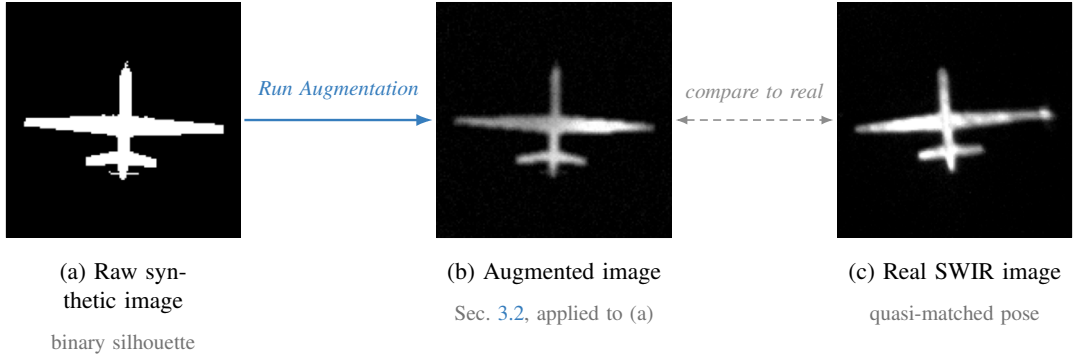

The model is trained on synthetic images and evaluated on real SWIR imagery, so its real-world accuracy depends on how closely the synthetic \emph{appearance} matches the sensor. Each synthetic training image is a binary silhouette mask: a uniform white aircraft on a black background. A real SWIR image of the same aircraft is far from binary, so the gap to be closed is one of \emph{intra-silhouette appearance}, meaning how the sensor fills in and blurs a shape, rather than of texture or lighting. Closing it therefore does not require a photorealistic renderer, only a statistical model of what the sensor does to a silhouette. Three such models are compared in this work; they are described below and summarized in \cref{tab:augmodels}, and \cref{fig:augpanel} shows the effect of the one adopted.

\paragraph{Measured target statistics.} The target appearance is characterized by pixel statistics measured on a single real image, using only raw intensities and never its pose label (the standard unsupervised-adaptation setting). On a $[0,1]$ intensity scale, the aircraft's interior has mean $0.537$ and standard deviation $0.172$, a wide and non-uniform distribution rather than a uniform plate. About $44\%$ of the pixels that a loose threshold assigns to the aircraft in fact belong to a dim \emph{halo} of intensity $0.10$--$0.20$ just outside its true boundary. The background is low but nonzero, with mean $0.0008$. Finally, the silhouette edge is blurred with an effective Gaussian scale of $\sigma\approx1$--$3$\,px, estimated from the $10$--$90\%$ rise width of the edge profile.

\paragraph{Three appearance models.} The three models differ only in how the binary mask is turned into a training image; the rest of training is identical. Each is sampled independently for every training image. Models (ii) and (iii) both begin with the same small perturbation of the silhouette boundary: with probability $0.1$ the mask is dilated or eroded, with equal chance, by one pixel using a $2{\times}2$ kernel. Global contrast $c$ and brightness $b$ are applied as $I \mapsto (I-0.5)\,c + 0.5 + b$, pivoting about mid-grey.

\emph{(i) None.} The binary mask is used directly, so the aircraft is a uniform plate of intensity $1$ on a background of $0$.

\emph{(ii) Flat-plate-plus-blur.} This is a conventional photometric augmentation~\cite{shorten2019survey}, of the form previously used in this pipeline. After the boundary perturbation, the mask is blurred with a Gaussian of $\sigma\in[0.30,0.85]$\,px, given a global contrast $c\in[0.88,1.08]$ and brightness $b\in[-0.010,0.020]$, and corrupted with independent per-pixel Gaussian noise of standard deviation drawn from $[0.002,0.018]$. The aircraft remains a uniform plate whose interior is changed only by the global contrast and the noise. Compared with the measured statistics, this leaves the aircraft about $3\times$ too bright (interior mean $\approx0.93$ against $0.54$), uses a blur $\sigma$ $4$--$10\times$ smaller than the measured edge scale, and adds background noise about $6.5\times$ the real level. More fundamentally, it cannot reproduce the real interior's large, spatially correlated variation, because noise drawn independently at each pixel is averaged away by any blur.

\emph{(iii) Calibrated (adopted).} Rather than perturbing a uniform plate, this model rebuilds the aircraft's intensity from three components, each with parameters drawn from a range centred on the measured values. The first is a \emph{core intensity field},
\begin{equation}
I_{\text{core}}(\mathbf{x}) = \big[\lambda_{\text{lo}} + (\lambda_{\text{hi}}-\lambda_{\text{lo}})\,N(\mathbf{x})\big]\,\min\!\big(1,\ \tau\,D(\mathbf{x})\big),
\label{eq:core}
\end{equation}
defined on the aircraft's interior. Here $N$ is Gaussian noise drawn on a grid $s$ times coarser than the image, with $s\in\{8,10,12,14\}$, then cubically upsampled and rescaled to $[0,1]$; drawing it coarsely keeps neighbouring pixels correlated, which is what reproduces the wide interior variation. $D$ is each interior pixel's distance to the silhouette boundary, divided by its maximum over the silhouette, so that $D=0$ at the edge and $D=1$ at the innermost point. The factor $\min(1,\tau D)$ therefore ramps the intensity up from zero at the edge, reaching its full value at a fraction $1/\tau$ of the way in. The ranges are $\lambda_{\text{lo}}\in[0.05,0.15]$, $\lambda_{\text{hi}}\in[0.95,1.10]$, and $\tau\in[1.2,1.8]$. The second component is an \emph{edge halo},
\begin{equation}
I_{\text{halo}}(\mathbf{x}) = \eta\,\max\!\big(0,\ 1 - d(\mathbf{x})/h\big),
\label{eq:halo}
\end{equation}
applied to the band of pixels outside the silhouette and within $h$ pixels of it, where $d$ is a pixel's distance to the silhouette, $h\in\{2,3,4\}$, and $\eta\in[0.08,0.14]$; it reproduces the measured dim halo. The third component is \emph{blur and background gating}. The combined image is blurred with a Gaussian of $\sigma\in[0.4,0.7]$\,px, and every pixel more than $h$ pixels from the silhouette is set to zero. Global contrast $c\in[0.90,1.08]$ and brightness $b\in[-0.01,0.01]$ are then applied; because they pivot about mid-grey, they would lift the zeroed background off the floor, so those background pixels are scaled back down by a factor of $0.02$ afterwards. Finally, per-pixel Gaussian noise of standard deviation drawn from $[5{\times}10^{-4},2.5{\times}10^{-3}]$ is added, matching the low real background level.

The calibrated model's Gaussian blur is no wider than that of model~(ii). The measured edge softness is reproduced instead by the interior taper of \cref{eq:core} and the exterior halo of \cref{eq:halo}, which together spread the transition from aircraft to background over several pixels; the final blur only smooths the boundaries between the components. This is the essential difference from model~(ii), which can soften an edge only by blurring and cannot change the uniform interior at all. The contribution of the calibrated model to accuracy is isolated by ablation in \cref{sec:sim2real-results}.

\begin{table}[htbp]
    \centering
    \footnotesize
    \setlength{\tabcolsep}{3pt}
    \caption{The three synthetic appearance models compared in this work, applied to the binary silhouette mask at training time. Ranges are sampled uniformly and independently for each image. Model~(iii) is used for all reported results; the three are compared in \cref{tab:sim2real}.}
    \label{tab:augmodels}
    \begin{tabular}{@{}lccc@{}}
        \toprule
        Step & (i) None & (ii) Flat-plate & (iii) Calibrated \\
        \midrule
        Boundary perturb. & -- & \checkmark & \checkmark \\
        Interior & uniform & uniform & \cref{eq:core} \\
        Edge halo & -- & -- & \cref{eq:halo} \\
        Blur $\sigma$ (px) & -- & $[0.30,0.85]$ & $[0.4,0.7]$ \\
        Contrast $c$ & -- & $[0.88,1.08]$ & $[0.90,1.08]$ \\
        Brightness $b$ & -- & $[-0.01,0.02]$ & $[-0.01,0.01]$ \\
        Bkg.\ gating & -- & -- & \checkmark \\
        Noise std & -- & $[0.002,0.018]$ & $[0.0005,0.0025]$ \\
        \bottomrule
    \end{tabular}
\end{table}

\subsection{Inference-Time Radar-Cue Branch Correction}
\label{sec:inference}

At inference the network produces a raw prediction $\R_{\text{pred}}$ lying on some member of the orbit. Its cue bits $(\widehat{\nts},\widehat{\tbs})$ are computed from \cref{eq:cues} and, given the reference bits $(\nts^\star,\tbs^\star)$ supplied by the radar signal, the correction emits the unique orbit member whose cues match:
\begin{equation}
    \R_{\text{out}} = \R_{\text{pred}}^{(i^\star)}, \quad
    i^\star \ \text{s.t.}\ (\nts,\tbs)\big(\R_{\text{pred}}^{(i^\star)}\big) = (\nts^\star,\tbs^\star).
    \label{eq:select}
\end{equation}
Because the two bits label the four orbit members distinctly (\cref{sec:ambiguity}), $i^\star$ is unique for nondegenerate poses; at the zero-crossing configurations discussed in (\cref{sec:results-main}), the cue may not uniquely distinguish the branches. Lacking a physical radar on the testbed, the reference bits are simulated by evaluating \cref{eq:cues}, exactly the range-rate and elevation signs a radar reports, and expose only those two bits to the correction, following~\cite{herrera2025}. This is \emph{not} an oracle that selects the lowest-error branch: it cannot recover the pose when the raw prediction is wrong on all four branches, and those failures populate the residual tail analyzed in \cref{sec:results}.
\section{Experimental Setup}
\label{sec:setup}

\subsection{Dataset}
\label{sec:realdata}

\begin{figure}[tb]
    \centering
    \begin{minipage}[b]{0.47\linewidth}
        \centering
        \includegraphics[width=\linewidth]{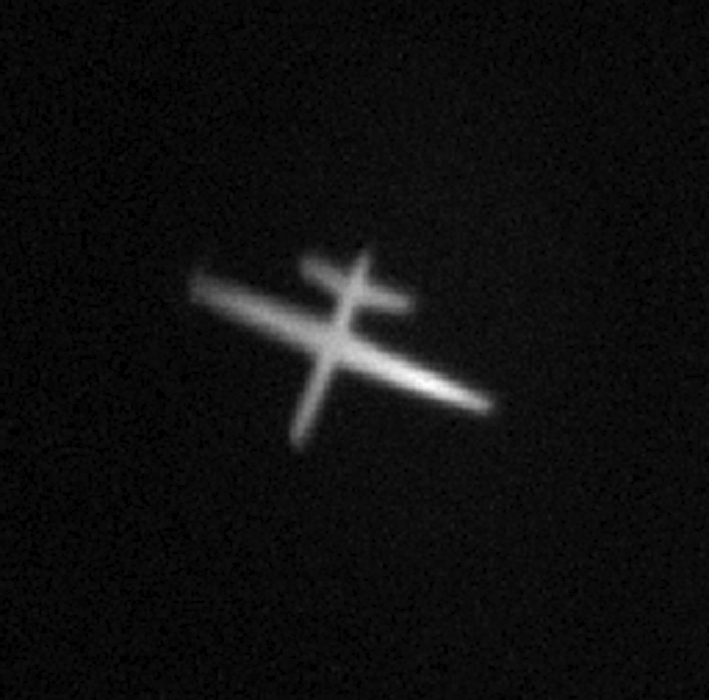}\\[2pt]
        {\footnotesize (a) real SWIR image}
    \end{minipage}\hfill
    \begin{minipage}[b]{0.47\linewidth}
        \centering
        \includegraphics[width=\linewidth]{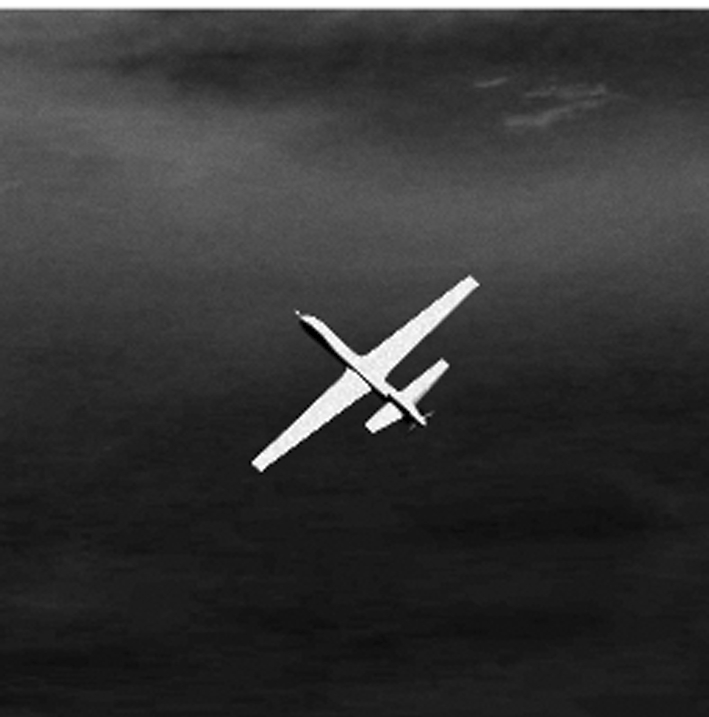}\\[2pt]
        {\footnotesize (b) synthetic sample}
    \end{minipage}
    \caption{A real SWIR sensor image (a) versus a synthetic sample (b), both from~\cite{herrera2025}. The gap the calibrated appearance model of \cref{sec:sim2real} must close is visible directly: the real target has a dim, non-uniform, softly blurred interior over a low but nonzero background, rather than the bright, flat, hard-edged silhouette a naive render produces. The measured statistics driving the calibration (foreground mean $0.537$, a dim edge halo, $1$--$3$px edge blur) are read from images of this type.}
    \label{fig:swir}
\end{figure}

The real dataset contains $77{,}077$ labeled SWIR images collected using the testbed described in~\cite{herrera2022aai}. A 3D-printed fixed-wing UAV model is mounted on a three-axis gimbal and imaged with a SWIR tracking camera. The gimbal motors are stepped through a dense grid of angular positions, and each image is assigned an orientation from the commanded motor positions. This provides hardware-captured images with precise ground truth. Further details on the testbed, motor position grid, and labeling procedure are given in~\cite{herrera2022aai}. \Cref{fig:swir} shows an example real image alongside a synthetic sample.
\subsection{Synthetic data and the synthetic-to-real protocol}
\label{sec:data}

The task is intrinsically a synthetic-to-real transfer problem: labels are cheap and exact in simulation but costly to obtain at scale on hardware, so the model is trained purely on synthetic renderings and evaluated purely on real sensor imagery. Training uses $100{,}000$ synthetic images (binary silhouette masks rendered from the target geometry over a dense grid of orientations), randomly partitioned into training, validation, and test sets in the proportions $80/10/10$. Ground-truth orientations are provided as axis-angle vectors (\cref{sec:rot}), and the images are made sensor-realistic by the calibrated appearance model of \cref{sec:sim2real}. Evaluation uses the $77{,}077$ real SWIR images of \cref{sec:realdata}. No real image is ever seen in training; every reported real-image number is a measurement of sim-to-real generalization.

\paragraph{Image size.}
Two image sizes are used in this work. The first is the size of the region cropped around the UAV, which determines how much background is included. Each real evaluation image is a $200{\times}200$-pixel crop from the $640{\times}513$ camera frame. Each synthetic image is cropped individually according to the size of the UAV in that image: the crop is a square centred on the UAV's bounding box, with a side length equal to $2.4$ times the longer side of the box. Because the bounding box changes with the UAV's orientation, the crop size in pixels varies from image to image, but the UAV always occupies the same fraction of the crop, its longer dimension spanning about $42\%$ ($1/2.4$) of the crop side. To guard against degenerate bounding boxes, the crop side is limited to between $12\%$ and $100\%$ of the longer side of the rendered image, and any part of the crop extending beyond the image border is filled with zeros. The second image size is the network input, which is fixed at $336{\times}336$ pixels because DINOv2 uses $14{\times}14$-pixel patches (\cref{sec:arch}). All real and synthetic crops are therefore resized to $336{\times}336$ using area-based interpolation before being passed to the network. For the $200{\times}200$ real images, this corresponds to an enlargement factor of $1.68$ and does not add new image detail; it simply matches the input resolution required by the backbone.

\subsection{Implementation details}
\label{sec:impl}
We use AdamW with a batch size of $32$ and train the model in two phases. In phase~1, the backbone is frozen and only the decoder and heads are trained for two epochs with a learning rate of $10^{-3}$. The chirality term is disabled during this phase because the frozen self-supervised features are nearly identical for an image and its mirror, making the chirality bit difficult to learn.
In phase~2, all $24$ backbone blocks are unfrozen and the chirality term is enabled. The backbone and decoder use separate learning rates of $3{\times}10^{-6}$ and $2{\times}10^{-4}$, respectively. The smaller learning rate for the backbone allows it to adapt toward mirror-sensitive features while preserving the pretrained representation. Both learning rates follow the same schedule: they increase linearly to their full values over the first three epochs, remain constant through epoch $110$, and then decay according to a cosine schedule toward $10\%$ of their initial values, reached at epoch $150$.

Training runs for up to $150$ epochs, with early stopping if the synthetic validation loss does not improve for $20$ consecutive epochs. The reported model is the checkpoint with the lowest synthetic validation loss, which occurs at epoch $102$. Real images are not used either for training or for checkpoint selection. The resulting model has $441.4$M parameters and is trained on a single NVIDIA L40S GPU. During evaluation, the $77{,}077$ real images are distributed across eight GPUs and processed in parallel, reducing evaluation time without changing the results.

\subsection{Evaluation protocol}
\label{sec:protocol}

For each real image, we compute the pose prediction and the radar-cue-corrected prediction described in \cref{sec:inference}. The geodesic error of each prediction relative to the ground-truth orientation is then computed using \cref{eq:geodesic}. We report the mean and median geodesic error, along with the fraction of predictions with errors greater than $90^\circ$. The mean and median describe the overall and typical prediction errors, respectively, while the $90^\circ$ threshold captures severe failures in which the estimated orientation is far from the ground truth. The evaluation is performed on all $77{,}077$ real SWIR images using $200{\times}200$ crops.
\section{Results and Discussion}
\label{sec:results}

\subsection{Dataset augmentation ablation}
\label{sec:sim2real-results}

\Cref{tab:sim2real} isolates the augmentation of \cref{sec:sim2real}: three training runs at $200{\times}200$ that differ only in how the synthetic silhouette is rendered (raw binary mask, generic flat-plate-plus-blur, and the calibrated model), with everything else fixed. Generic augmentation cuts mean error from $34.3^\circ$ to $28.4^\circ$; the calibration removes a further $9.3^\circ$, a larger step than adding augmentation at all, for a $44\%$ total reduction. The benefit comes from matching the synthetic appearance to measured sensor statistics, not from augmentation in the abstract.

\begin{table}[tb]
    \centering
    \small
    \setlength{\tabcolsep}{6pt}
    \caption{Augmentation ablation ($200{\times}200$, $77{,}077$ real images). The three runs differ only in how the synthetic silhouette is rendered during training.}
    \label{tab:sim2real}
    \begin{tabular}{@{}lccc@{}}
        \toprule
        Synthetic appearance model & Mean$^\circ$ & Med.$^\circ$ & RMSE$^\circ$ \\
        \midrule
        None (raw binary mask)      & 34.27 & 19.62 & 59.89 \\
        Flat-plate-plus-blur        & 28.44 & 14.81 & 47.52 \\
        \textbf{Calibrated (ours)}  & \textbf{19.18} & \textbf{10.58} & \textbf{38.42} \\
        \bottomrule
    \end{tabular}
\end{table}

\begin{figure*}[t]
    \centering

    \begin{subfigure}[b]{0.495\textwidth}
        \centering
        \includegraphics[width=\linewidth]{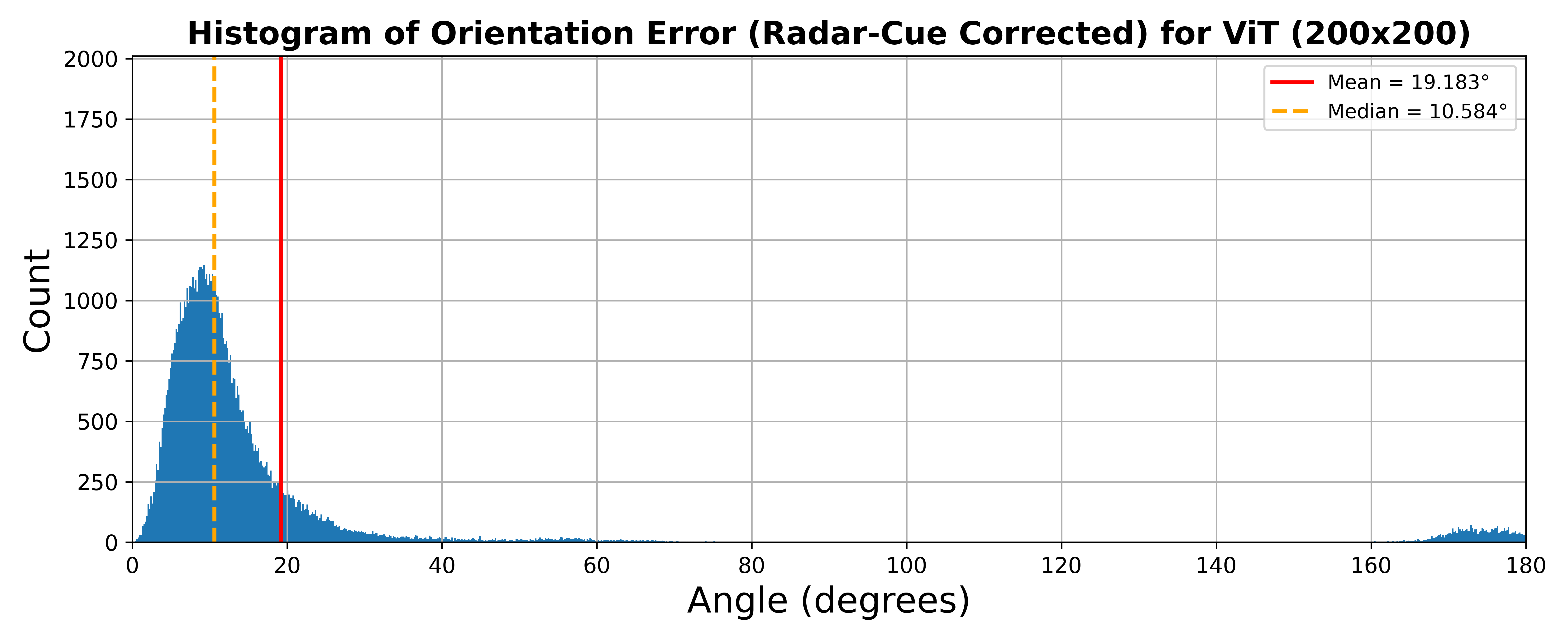}
        \caption{Single frame (mean $19.18^\circ$, median $10.58^\circ$).}
        \label{fig:raw-hist}
    \end{subfigure}%
    \hfill%
    \begin{subfigure}[b]{0.495\textwidth}
        \centering
        \includegraphics[width=\linewidth]{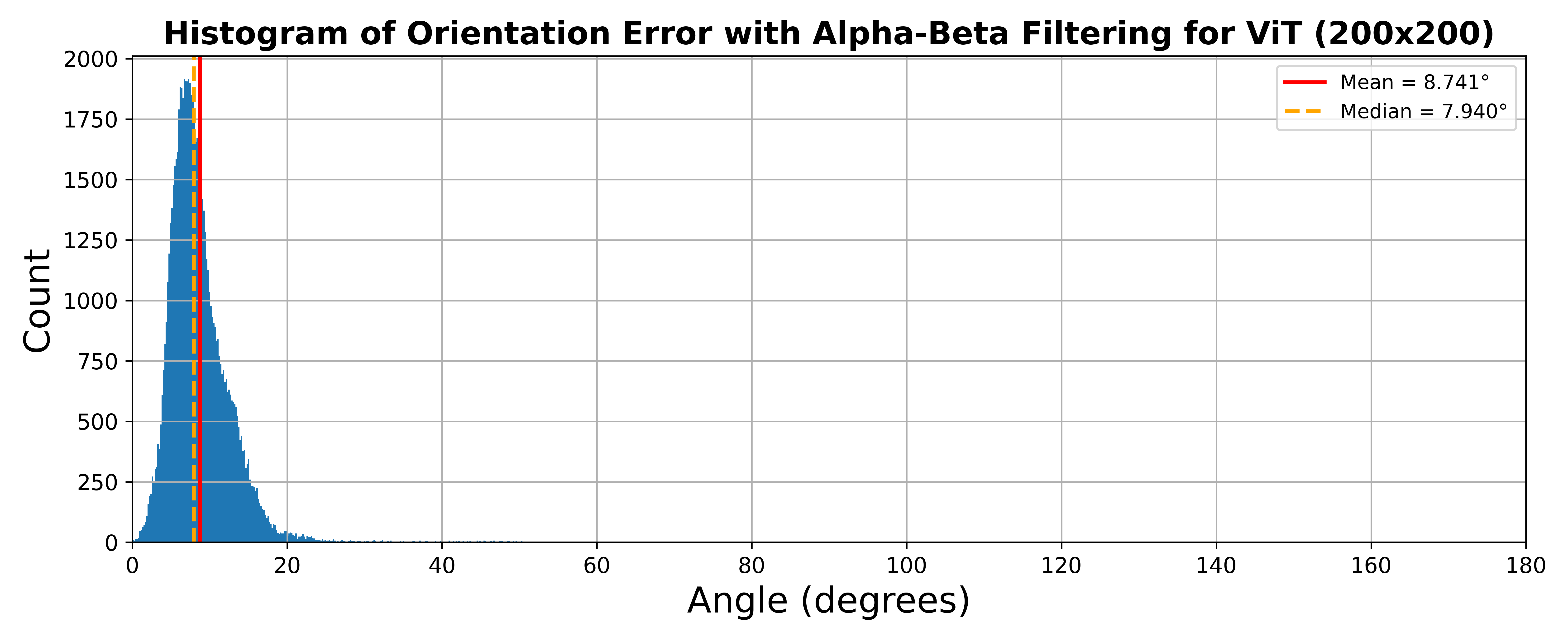}
        \caption{After $\alpha$-$\beta$ refinement (mean $8.74^\circ$, median $7.94^\circ$).}
        \label{fig:hist-compare}
    \end{subfigure}

    \caption{Error distribution at $200{\times}200$. (a) A single well-resolved mode near the median sits beside an isolated tail near $180^\circ$ from cases the radar cue cannot fix. (b) After the sequential $\alpha$-$\beta$ filter, the $180^\circ$ tail is gone and the main mode tightens around its median.}
    \label{fig:hists}
\end{figure*}

\subsection{Ambiguity resolution}
\label{sec:ambiguity-results}

\Cref{tab:ambiguity} compares the raw prediction against the radar-cue-corrected prediction on the same model and images. The raw median error is $167.6^\circ$ because the network lands on the correct one of the four mirror-related candidates only $18.1\%$ of the time. This measures the objective's indifference rather than any limit on what a frame could reveal: \cref{eq:symloss} exerts no pressure toward the true candidate by construction. That $18.1\%$ falls below the $25\%$ of a random assignment indicates a systematic preference among candidates, whose source is left to future work. This shows how the two stages divide the work: the radar cue picks which of the four candidates is correct, and the network's own accuracy determines how close the prediction lands within that candidate, reflected in the $10.58^\circ$ median. The correction isn't perfect: $4.1\%$ of predictions still end up on the wrong candidate.

\begin{table}[tb]
    \centering
    \small
    \setlength{\tabcolsep}{6pt}
    \caption{Effect of radar-cue branch selection ($200{\times}200$, $77{,}077$ real images). ``Flipped'' is the fraction of predictions on a mirror-related candidate other than the true one; the $4.1\%$ that survive the correction are the near-$180^\circ$ tail of \cref{fig:raw-hist}, explained in \cref{sec:results-main}.}
    \label{tab:ambiguity}
    \begin{tabular}{@{}lccc@{}}
        \toprule
        Prediction & Mean$^\circ$ & Med.$^\circ$ & Flipped \\
        \midrule
        Raw (no correction)          & 130.51 & 167.57 & 81.9\% \\
        \textbf{Radar-cue corrected} & \textbf{19.18} & \textbf{10.58} & 4.1\% \\
        \bottomrule
    \end{tabular}
\end{table}

\subsection{Single-frame pose estimation}
\label{sec:results-main}

\Cref{tab:main} reports the model on the $77{,}077$ real images at the $200{\times}200$ condition, both as a single-frame estimate after radar-cue correction and after the sequential $\alpha$-$\beta$ refinement of \cref{sec:residual-refinement}. From a single frame the model attains $19.18^\circ$ mean and $10.58^\circ$ median geodesic error ($38.42^\circ$ RMSE), with $45.4\%$ of predictions inside $10^\circ$ and $4.1\%$ beyond $90^\circ$. \Cref{fig:hists} plots the error distribution before and after temporal refinement. Before refinement (\cref{fig:raw-hist}) a single well-resolved mode near the median sits beside a small, isolated tail at $\sim\!180^\circ$, with almost no predictions falling in between, as the histogram shows.

\begin{figure*}[t]
    \centering
    \begin{minipage}[t]{0.49\textwidth}
        \begin{tcolorbox}[
            colback=green!3,
            colframe=green!55!black,
            arc=4mm,
            boxrule=1pt,
            left=2pt, right=2pt, top=2pt, bottom=2pt,
            title=\textbf{Accurate predictions},
            fonttitle=\bfseries,
            coltitle=green!35!black,
            colbacktitle=green!15,
            attach boxed title to top center={yshift=-2mm},
            boxed title style={colframe=green!55!black, arc=2mm, boxrule=0.6pt}
        ]
            \centering
            \begin{subfigure}[b]{0.48\linewidth}
                \centering
                \includegraphics[width=0.92\linewidth]{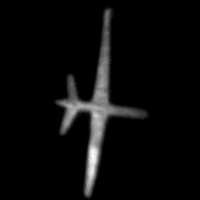}
                \caption*{\scriptsize Angle Error $= 0.24^\circ$}
            \end{subfigure}\hfill
            \begin{subfigure}[b]{0.48\linewidth}
                \centering
                \includegraphics[width=0.92\linewidth]{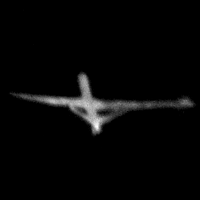}
                \caption*{\scriptsize Angle Error $= 0.53^\circ$}
            \end{subfigure}
        \end{tcolorbox}
    \end{minipage}
    \hfill
    \begin{minipage}[t]{0.49\textwidth}
        \begin{tcolorbox}[
            colback=red!3,
            colframe=red!65!black,
            arc=4mm,
            boxrule=1pt,
            left=2pt, right=2pt, top=2pt, bottom=2pt,
            title=\textbf{Catastrophic predictions},
            fonttitle=\bfseries,
            coltitle=red!55!black,
            colbacktitle=red!15,
            attach boxed title to top center={yshift=-2mm},
            boxed title style={colframe=red!65!black, arc=2mm, boxrule=0.6pt}
        ]
            \centering
            \begin{subfigure}[b]{0.48\linewidth}
                \centering
                \includegraphics[width=0.92\linewidth]{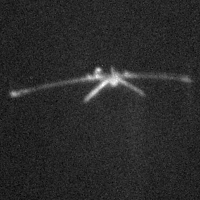}
                \caption*{\scriptsize Angle Error $= 179.98^\circ$}
            \end{subfigure}\hfill
            \begin{subfigure}[b]{0.48\linewidth}
                \centering
                \includegraphics[width=0.92\linewidth]{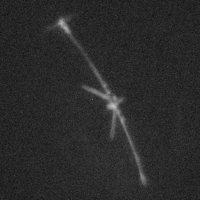}
                \caption*{\scriptsize Angle Error $= 180.00^\circ$}
            \end{subfigure}
        \end{tcolorbox}
    \end{minipage}
    \caption{Representative accurate and catastrophic radar-cue-corrected predictions from the ViT $200\times200$ condition. Accurate cases (left, green) are viewed near broadside, where illumination strikes the airframe close to perpendicular and produces uniform, well-resolved reflectance. Catastrophic cases (right, red) are viewed at high tilt, near the edge-on geometry, where oblique illumination produces nonuniform reflectance across the airframe and the geodesic cue becomes least reliable at exactly the geometry where it is most needed.}
    \label{fig:qualitative-cases}
\end{figure*}

\paragraph{What remains at $180^\circ$.} If the two bits resolve the nose--tail and top--bottom ambiguities, why does any error near $180^\circ$ survive? The bits are never undefined: by \cref{eq:cues}, nts and tbs always take a value in $\{+1,-1\}$. What fails is reliability, not definition. Since $(\mathbf{R}^{(3)}\mathbf{e}_3)_z = -(\mathbf{R}\mathbf{e}_3)_z$, the true pose and its top--bottom flip carry opposite $\mathrm{tbs}$ bits everywhere except the single point $(\mathbf{R}\mathbf{e}_3)_z = 0$; near that point the separation is small enough that ordinary prediction error flips it, and picking wrong gives exactly $180^\circ$, since the two candidates differ by a rotation of $\pi$ about the body axis. This is physical rather than numerical: $(\mathbf{R}\mathbf{e}_1)_z \approx 0$ is the broadside geometry where a radar measures zero range rate and genuinely cannot say whether the nose approaches, and $(\mathbf{R}\mathbf{e}_3)_z \approx 0$ is the edge-on geometry where upright and inverted are indistinguishable. The cue is unreliable there for the same reason the image is. Consistent with this, the two failures in \cref{fig:qualitative-cases} have errors of $179.98^\circ$ and $180.00^\circ$, the signature of a wrong candidate rather than a poor regression, and one sits at $(\mathbf{R}\mathbf{e}_3)_z \approx 0.03$.

\begin{table}[tb]
    \centering
    \small
    \setlength{\tabcolsep}{6pt}
    \caption{Geodesic error on $77{,}077$ real images (DINOv2-Large, epoch $102$, $200{\times}200$ crops), from a single frame and after the sequential $\alpha$-$\beta$ filter of \cref{sec:residual-refinement}.}
    \label{tab:main}
    \begin{tabular}{@{}lccc@{}}
        \toprule
        Estimate ($200{\times}200$) & Mean$^\circ$ & Med.$^\circ$ & $>\!90^\circ$ \\
        \midrule
        Single frame, radar-cue corrected & 19.18 & 10.58 & 4.1\% \\
        $+$ sequential $\alpha$-$\beta$ refinement & \textbf{8.74} & \textbf{7.94} & \textbf{0.0\%} \\
        \bottomrule
    \end{tabular}
\end{table}

\subsection{Sequential \texorpdfstring{$\alpha$-$\beta$}{alpha-beta} refinement}
\label{sec:residual-refinement}

The $4.1\%$ of single-frame predictions beyond $90^\circ$ are the residual mirror flips the radar cue cannot fix. Because the real images are captured as a raster scan of the gimbal motors in $3^\circ$ steps, consecutive frames are close in orientation, which a constant-rate temporal filter can exploit to suppress isolated errors; this is also representative of deployment, where a maneuvering target traces a continuous attitude trajectory. The $\alpha$-$\beta$ filter of~\cite{herrera2025} is adopted and applied as a post-hoc smoother with an outlier gate on the radar-cue-corrected $R_{\text{out}}$ of \cref{eq:select}.

\begin{figure}[tb]
\centering
\usetikzlibrary{arrows.meta,positioning,calc,shapes.geometric}
\definecolor{wacvblue}{RGB}{28,86,150}
\definecolor{kinorange}{RGB}{201,120,32}
\definecolor{coastred}{RGB}{176,58,58}
\definecolor{updgreen}{RGB}{42,120,66}
\definecolor{fnneutral}{RGB}{70,74,82}
\providecommand{\absub}{\scriptsize\color{black!60}}
\resizebox{\columnwidth}{!}{%
\begin{tikzpicture}[
    font=\small,
    >=Latex,
    box/.style={draw, semithick, rounded corners=2.5pt, align=center,
                minimum height=0.82cm, inner sep=3pt, text=black!88},
    meas/.style ={box, draw=wacvblue,   fill=wacvblue!8,   minimum width=3.60cm},
    kin/.style  ={box, draw=kinorange,  fill=kinorange!10, minimum width=3.60cm},
    calc/.style ={box, draw=fnneutral!70,fill=fnneutral!7, minimum width=4.10cm},
    coast/.style={box, draw=coastred,   fill=coastred!7,   minimum width=3.45cm},
    upd/.style  ={box, draw=updgreen,   fill=updgreen!7,   minimum width=3.45cm},
    gate/.style ={draw, semithick, diamond, aspect=2.6, draw=fnneutral!70,
                  fill=fnneutral!7, align=center, inner sep=1pt, text=black!88},
    arr/.style={->, semithick, draw=fnneutral!75, rounded corners=2pt},
    lbl/.style={font=\scriptsize\bfseries, inner sep=2pt}
]

\node[meas] (meas) at (-1.92,0)
    {Model prediction $R_k$\\[1pt]{\absub radar-cue corrected, \cref{eq:select}}};
\node[kin]  (pred) at ( 1.92,0)
    {Kinematics pred.\ $\hat{R}^{p}_{k}$\\[1pt]{\absub propagated from $k{-}1$}};

\node[calc] (residual) at (0,-1.60)
    {Residual $y_k$\\[1pt]{\absub \cref{eq:residual}}};

\node[gate] (gate) at (0,-2.95) {$\|y_k\| < \tau$\,?};

\node[coast] (no)  at (-2.05,-4.30)
    {Coast on kinematics\\[1pt]{\absub reject measurement}};
\node[upd]   (yes) at ( 2.05,-4.30)
    {Update pose and rate\\[1pt]{\absub accept measurement}};

\node[calc] (out) at (0,-5.85)
    {Filtered estimate $\hat{R}^{u}_{k}$\\[1pt]{\absub output for sample $k$}};

\draw[arr, draw=wacvblue]  (meas.south) -- ++(0,-0.45) -| ([xshift=-0.9cm]residual.north);
\draw[arr, draw=kinorange] (pred.south) -- ++(0,-0.45) -| ([xshift= 0.9cm]residual.north);
\draw[arr] (residual) -- (gate);

\draw[arr, draw=coastred] (gate.west)
    -| node[lbl, above, pos=0.30, text=coastred] {no} (no.north);
\draw[arr, draw=updgreen] (gate.east)
    -| node[lbl, above, pos=0.30, text=updgreen] {yes} (yes.north);

\draw[arr, draw=coastred] (no.south)  -- ++(0,-0.45) -| ([xshift=-0.9cm]out.north);
\draw[arr, draw=updgreen] (yes.south) -- ++(0,-0.45) -| ([xshift= 0.9cm]out.north);

\draw[arr, draw=kinorange, dashed, dash pattern=on 2pt off 1.6pt]
    (out.east) -- (4.10,-5.85)
    -- (4.10,0) -- (pred.east);
\node[font=\scriptsize\itshape, text=kinorange, align=center, anchor=east]
    at (4.1,-2.2) {propagate\\to sample\\$k{+}1$};

\end{tikzpicture}%
}
\caption{The predict-gate-update cycle of the $\alpha$-$\beta$ filter (\crefrange{eq:predict}{eq:update}), with equations adopted from~\cite{herrera2025}. A radar-cue-corrected prediction updates a running pose/rate estimate only when its residual falls below threshold $\tau$; otherwise the estimate coasts on kinematics alone.}
\label{fig:abfilter}
\end{figure}
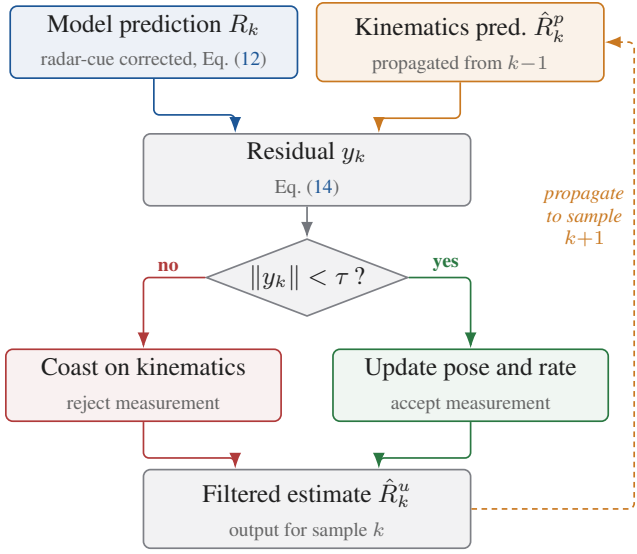

\paragraph{Filter formulation.} Writing $R_k$ for the radar-cue-corrected prediction at sample $k$ and $(\hat{R}_k^u, \hat{\omega}_k^u)$ for the filter's updated pose and angular-velocity estimate, the filter predicts, computes a residual, and conditionally updates:
\begin{align}
\hat{R}_k^p &= \exp\left(\Delta T \hat{\omega}_{k-1}^u\right) \hat{R}_{k-1}^u, \quad \hat{\omega}_k^p = \hat{\omega}_{k-1}^u \label{eq:predict}\\[4pt]
y_k &= \log\left(R_k \left(\hat{R}_k^p\right)^T\right) \label{eq:residual}\\[4pt]
\left(\hat{R}_k^u, \hat{\omega}_k^u\right) &=
\begin{cases}
\left(\exp(\alpha y_k) \hat{R}_k^p,\ \hat{\omega}_k^p + \frac{\beta}{\Delta T} y_k\right) & \|y_k\| < \tau \\[10pt]
\left(\hat{R}_k^p,\ \hat{\omega}_k^p\right) & \text{otherwise}
\end{cases} \label{eq:update}
\end{align}
where $\alpha, \beta$ are the filter gains and $\tau$ is an outlier-rejection threshold: a measurement updates the estimate only if its residual falls below $\tau$; otherwise the filter coasts on kinematics alone. Values used were $\alpha = 0.0356$, $\beta = 0.00132$, $\tau = 137.3^\circ$, and $\Delta T = 1$.
\Cref{fig:abfilter} illustrates this predict-gate-update cycle.

\subsection{Inference latency}
\label{sec:latency}

Because frames are processed sequentially in the UAV application, single-image latency is reported on one L40S at $200{\times}200$ (\cref{tab:speed}), as mean $\pm$ std over five independent job submissions of 200 timed forward passes each. The forward pass takes $18.02\pm0.21$~ms in fp32 and $13.25\pm0.12$~ms under fp16 with \texttt{torch.compile}, roughly 75 frames per second; the $\alpha$-$\beta$ filter adds under 0.17~ms.

\begin{table}[tb]
\centering
\small
\setlength{\tabcolsep}{7pt}
\caption{Single-image inference latency at $200{\times}200$ on one L40S, forward pass alone and with the $\alpha$-$\beta$ filter.}
\label{tab:speed}
\begin{tabular}{@{}lcc@{}}
\toprule
Precision & Model (ms) & $+$Filter (ms) \\
\midrule
fp32            & $18.02 \pm 0.21$ & 18.19 \\
fp16, compiled  & $13.25 \pm 0.12$ & 13.42 \\
\bottomrule
\end{tabular}
\end{table}

\subsection{Qualitative failure analysis}
\label{sec:qualitative}

\Cref{fig:qualitative-cases} shows four representative frames: two accurate predictions (sub-degree error) and two catastrophic failures ($\sim\!180^\circ$). Two factors distinguish them. First, the failures sit near the degenerate viewing configuration where $(\mathbf{R}\mathbf{e}_1)_z$ or $(\mathbf{R}\mathbf{e}_3)_z$ approaches zero: the fuselage is heavily foreshortened, the wings and tail collapse toward the same silhouette under the reflections of \cref{eq:branches}, and the geometric cue of \cref{eq:cue} becomes noise-dominated because the fin offset $p$ approaches zero. The accurate cases are viewed near broadside or at moderate angles where all parts are separately resolvable. Second, the failures are viewed at higher tilt, where illumination strikes the airframe obliquely rather than perpendicular to its surface, producing nonuniform reflectance across the fuselage; the accurate cases, viewed closer to broadside, receive more uniform illumination and present cleaner features to the network.

\section{Conclusion}
\label{sec:conclusion}
This work investigated the synthetic-to-real generalization capability of a ViT-based model for monocular pose estimation of a noncooperative UAV. Using a self-supervised DINOv2 backbone, the model was trained exclusively on labeled synthetic imagery and evaluated on 77,077 labeled real-world UAV images. Pose ambiguity-aware strategies were incorporated during training and inference, and an $\alpha$--$\beta$ filter was used to further improve pose estimates. The model achieved a mean angular error of $19.18^{o}$ with an inference time of $13.25$ ms before filtering, and $8.74^{o}$ and $13.42$ ms, respectively, after filtering. 

The mean angular error of $8.74^{o}$ is considered acceptable given three identified factors that contribute to this error: the synthetic-to-real domain gap, pose misalignment in the real-world dataset resulting from experimental imperfections, and the inherent prediction error of deep-learning models. Considering these factors, the results demonstrate the potential of the proposed model for real-world pose estimation of noncooperative UAVs.

Future work will investigate further domain-gap reduction techniques, including the use of photogrammetry to reconstruct a 3D virtual model directly from a real UAV. This approach could enable the generation of virtually unlimited labeled training imagery that retains the visual characteristics of the real UAV, thereby useful to reduce the synthetic-to-real domain gap.

{\small
\bibliographystyle{ieeenat_fullname}
\bibliography{main}
}
\end{document}